\documentclass[12pt]{article}

\usepackage{newtxtext,newtxmath}

\usepackage{graphicx}

\usepackage[letterpaper,margin=1in]{geometry}

\renewenvironment{abstract}
	{\quotation}
	{\endquotation}

\date{}

\makeatletter
\renewcommand{\fnum@figure}{\textbf{Figure \thefigure}}
\renewcommand{\fnum@table}{\textbf{Table \thetable}}
\makeatother

\usepackage{scicite}

\usepackage{url}

\def\scititle{
	Proposing and solving olympiad geometry with guided tree search
}
\title{\bfseries \boldmath \scititle}

\author{
	Chi~Zhang$^{1\ast}$ \and Jiajun~Song$^{1}$ \and Siyu~Li$^{1,2}$ \and Yitao~Liang$^{1,2}$ \and Yuxi~Ma$^{1,2}$ \and Wei~Wang$^{1}$ \and Yixin~Zhu$^{2\ast}$ \and Song-Chun~Zhu$^{1,2\ast}$ \and
	\small$^{1}$Beijing Institute for General Artificial Intelligence (BIGAI), Beijing, 100080, China.\and
	\small$^{2}$Institute for Artificial Intelligence, Peking University, Beijing, 100080, China.\and
	\small$^\ast$Corresponding author. Email: zhangchi@bigai.ai, yixin.zhu@pku.edu.cn, sczhu@bigai.ai
}

\begin{document}

\maketitle

\begin{abstract} \bfseries \boldmath
Mathematics olympiads are prestigious competitions, with problem proposing and solving highly honored. Building artificial intelligence that proposes and solves olympiads presents an unresolved challenge in automated theorem discovery and proving~\cite{selsam2020imo}, especially in geometry for its combination of numerical and spatial elements. We introduce TongGeometry, a Euclidean geometry system supporting tree-search-based guided problem proposing and solving. The efficient geometry system establishes the most extensive repository of geometry theorems to date: within the same computational budget as the existing state-of-the-art~\cite{trinh2024solving}, TongGeometry discovers 6.7 billion geometry theorems requiring auxiliary constructions, including 4.1 billion exhibiting geometric symmetry. Among them, 10 theorems were proposed to regional mathematical olympiads with 3 of TongGeometry's proposals selected in real competitions, earning spots in a national team qualifying exam or a top civil olympiad in China and the US. Guided by fine-tuned large language models, TongGeometry solved all International Mathematical Olympiad geometry in IMO-AG-30, outperforming gold medalists for the first time. It also surpasses the existing state-of-the-art across a broader spectrum of olympiad-level problems. The full capabilities of the system can be utilized on a consumer-grade machine, making the model more accessible and fostering widespread democratization of its use. By analogy, unlike existing systems that merely solve problems like students, TongGeometry acts like a geometry coach, discovering, presenting, and proving theorems.
\end{abstract}


\paragraph{Significance} \textit{This research presents TongGeometry, a system for Euclidean geometry problem proposing and solving, bridging numerical and spatial reasoning. TongGeometry surpasses state-of-the-art methods by autonomously discovering 6.7 billion theorems requiring auxiliary constructions, including 4.1 billion exhibiting geometric symmetry. Its real-world impact is demonstrated by its proposed problems being featured in prestigious math olympiads and outperforming top human competitors in solving International Mathematical Olympiad geometry problems. This achievement represents a step towards both automated theorem discovery and problem-solving, akin to transitioning from a ``student'' system to a ``coach'' capable of both generating and solving advanced problems. Importantly, its efficiency and accessibility on consumer-grade machines democratize high-level geometric problem-solving, enabling broader educational and research applications.}

\paragraph{}

Theorem proving, the journey to navigate from the initial state to the goal by connecting potentially missing intermediate conditions, demonstrates exceptional problem-solving skills. For any individual, winning a prize in the International Mathematical Olympiad (IMO) is a distinct honor and a testament to his / her problem-solving abilities. Since the inception, IMO has become the premier platform for identifying the world's best mathematical talents. It has also, inadvertently, become the target for Artificial Intelligence (AI) researchers striving to develop theorem-proving models that can match or surpass human competitors in the most prestigious mathematical competition in the world~\cite{selsam2020imo,polu2020generative,zheng2021minif2f,polu2022formal,lample2022hypertree}.

Proposing olympiad problems is held in high regard within the mathematics community as well, yet few computational methods have been able to accomplish autonomous problem proposing and rigorous proposal verification. Olympiad problems are often founded on years of intermediate theorems discovered and come with stringent criteria. The most exemplary problems appear deceptively simple and are comprehensible with a basic level of mathematical knowledge but require exceptional ingenuity to solve completely. Additionally, elegance, such as various forms of symmetry, is typically essential.

Proposing or solving olympiad problems poses considerable challenges for computational methods due to the vast number of branching points and the limited pool of past problems available for heuristics development. Geometry, in particular, stands out among olympiad topics because it involves both numerical and spatial reasoning, unlike other subjects that are text-based and can be more easily adapted with pretrained Large Language Models (LLMs)~\cite{achiam2023gpt,touvron2023llama}. Besides, contentious axiomatization in Euclidean geometry invalidates either completeness~\cite{ball1960short} or human-readability of interactive proof assistants applicable in other domains~\cite{maric2020formalizing,moura2021lean}. Although specific languages tailored for geometry have been created~\cite{chou1996introduction,sangwin2007brief,bak2020automated,trinh2024solving}, their design limitations leave the space of olympiad problems inadequately explored.

In this work, we introduce TongGeometry\footnote{Tong stands for ``general'' in Chinese.}, a tree-based system for synthetic Euclidean planar geometry that facilitates human-readable problem proposing via backward tracing and theorem proving through forward chaining. Utilizing this scalable and efficient system and 196 olympiad problems as guiding statistics, we collected 4.1 billion symmetric geometry problems with proofs, among 6.7 billion that all necessitate auxiliary constructions for a complete proof. By analogy, compared to existing ``student'' systems that only solve problems~\cite{sangwin2007brief,bak2020automated,trinh2024solving}, TongGeometry serves as a ``coach'' that characterizes the space of geometry using a finite tree and enables guided problem proposing and proving. We also established scoring rubrics for the difficulty and suitability of problems for olympiad competitions and selected a few as proposals. One was selected in 2024 National High School Mathematics League (Beijing), a qualifying competition for Chinese National Team\footnote{Team China won the first place in IMO 2019-2023 and the second place in IMO 2024.}, and two made the shortlist of the 2024 US Ersatz Math Olympiad, a top civil mathematics olympiad in the US\footnote{The civil competition is organized by Team USA coaches. Team USA won the first place in IMO 2024.}.

The generated data contains abundant auxiliary constructions for solving geometry problems. Filling in these auxiliary constructions is crucial for successful geometry theorem proving; these exogenous objects enable a proving system to bridge the gap between the initial state and the goal. We therefore leveraged the data to guide TongGeometry tree search when it is presented a problem to solve. Specifically, we fine-tuned two LLMs~\cite{guo2024deepseek,hoffmann2022training}: one dedicated to suggesting possible search directions and another for estimating the number of steps to go in each direction. When tested on the past 23 years of IMO geometry problems in the IMO-AG-30 benchmark, the neuro-symbolic system solves each of them within 38 minutes on a consumer-grade machine with 32 CPU cores and a single GPU card, becoming the first to surpass an average IMO gold medalist~\cite{trinh2024solving}.

\section*{Assessing TongGeometry}

\subsection*{Evaluating generated problems for olympiads}

Using 196 existing olympiad problems as guiding statistics, we performed massive parallel problem search using 10,368 parallel CPU cores. In 30 days of search, TongGeometry traverses 143,379,886 unique paths (170,883,417 in total) in the defined space of geometry, inferring over 1,851,166,755 unique states. On each unique path, TongGeometry finds on average 0.7613 configuration requiring auxiliaries, resulting in 109,157,477 configurations (pairs of context and auxiliaries). Among them, 70,703,508 are unique. After filtering, we ended up with a dataset of 6,688,310,403 problems (triplets of context, goal and auxiliaries), of which 4,096,680,574 are symmetric. See Figure~\ref{fig:stats_examples}(B) for comparison of search efficiency with AlphaGeometry.

With billions of problems available, the challenge of selectively retrieving suitable ones for olympiad proposals remains unresolved due to the absence of an automatic assessment scheme. Inspired by GeoGen's methodology~\cite{bak2020automated}, we developed a set of rubrics for problem selection.

Before the problem proposal deadlines for the 2024 National High School Mathematics League (Beijing) and the 2024 US Ersatz Math Olympiad, we enlisted the expertise of a 2023 IMO gold medalist and a student member of Chinese National Team to manually examine a batch of proposals during the initial phase of problem search. 4 proposals were submitted to the 2024 National High School Mathematics League (Beijing), with 1 selected as the only geometry problem in the final competition. Additionally, 6 proposals were submitted to the 2024 US Ersatz Math Olympiad, with 2 making it to the shortlist (Figure~\ref{fig:stats_examples}(E) and Supplementary Text). Of note, AlphaGeometry solved 3 of our 10 proposals.

\subsection*{Search discovers known lemmas and base configurations}

Among the proposed problems, we discover well-known lemmas. Figure~\ref{fig:stats_examples}(D) shows an rotationally symmetric lemma known in the community, where the three green lines are concurrent at the Nine-point center~\cite{mackay1892history,chen2021euclidean}. See Supplementary Text for explanation.

To facilitate problem exploration, we implemented a replay buffer that allowed tree search to restart from cached promising intermediate states. These states include foundational configurations well-known in the community, which serve as scaffolds for constructing more complex problem scenarios. Figure~\ref{fig:stats_examples}(F) illustrates a symmetric example featuring the Mixtilinear Incircle~\cite{chen2015guessing} and the Incenter/Excenter Lemma~\cite{chen2016incenter}. This configuration has been acclaimed as the ``richest configuration'' by a coach of the USA team~\cite{chen2015guessing}. See Supplementary Text for explanation.

\subsection*{Performance on benchmarks}

We performed quantitative analysis of TongGeometry on two benchmarks, the IMO-AG-30 dataset that was curated in AlphaGeometry and the newly curated dataset in the development of TongGeometry coined MO-TG-225. The IMO-AG-30 dataset comprises 30 problems, derived by translating 23 years of IMO problems into the domain-specific language of AlphaGeometry. In contrast, the MO-TG-225 dataset includes 225 mathematical olympiad problems selected from our pool of 196 examples used to calculate search statistics. Problems in MO-TG-225 have been translated into the domain-specific language of TongGeometry, and none of these problems appear in TongGeometry's training dataset.

In experiments, we compared TongGeometry with AlphaGeometry, GPT-4~\cite{achiam2023gpt} and its o1 variant. Given that TongGeometry and AlphaGeometry employed distinct domain-specific languages, we translated each problem's original representation into the respective language required by each model. For evaluating GPT-4 and o1, we used the natural language format in which each problem was presented in the competition. All models were assessed with a standardized 90-minute time limit per problem.

Table~\ref{tab:perf}(left) shows the performance of different models and human averages on IMO-AG-30. While the latest reasoning-enhanced language model of o1 achieved impressive results on a range of tasks, we observed that large language models still struggle with rigorous mathematical reasoning in geometry, often generating proofs with erroneous logics and hallucinated intermediate results. AlphaGeometry's DD+AR approach notably improved upon Wu's method (10 solves), yet its symbolic reasoning engine remains both redundant and suboptimal. By redesigning and optimizing the deductive database method, TongGeometry's reasoning backend, using only DD, successfully solved 18 problems --- outperforming average IMO contestants. With a learned value heuristics, TongGeometry became the first method to surpass IMO gold medalists, successfully proving all 30 problems in the benchmark. A closer examination revealed that the value model was instrumental in solving the benchmark's two most challenging problems: IMO 2000 P6 and IMO 2008 P6. Compared to AlphaGeometry, TongGeometry achieved these results on a consumer-grade machine with 32 CPU cores and a single NVIDIA RTX 4090 GPU in a maximum of 38 minutes, whereas AlphaGeometry required 246 CPU cores and 4 NVIDIA V100 GPUs to reduce solve time to under 90 minutes --- a resource-intensive setup inaccessible to most users.

Table~\ref{tab:perf}(right) displays the performance of various models on the MO-TG-225 dataset. Unlike the IMO dataset, MO-TG-225 includes problems from diverse sources, making rigorous human evaluation benchmarks unavailable. During the evaluation of GPT-4 and o1, we observed that when prompted to prove a known theorem, such as the Euler line theorem, LLMs frequently assumed the theorem as established and applied it directly without the requested proof. We considered these responses as correct, accounting for a few solves reported in the table. Consistent with results on IMO-AG-30, TongGeometry's DD backend demonstrated improved problem-solving capability over AlphaGeometry's DD+AR, reaching a performance level close to AlphaGeometry overall. We noted that AlphaGeometry's success largely stemmed from its backend engine, with 72.5\% of its total solves achieved by DD+AR. In contrast, TongGeometry not only solved a greater proportion of problems (81.3\% vs. 45.3\%) but also leveraged its neural models to address auxiliary construction effectively, with only 55.2\% solved by DD.

Ablation study on the value heuristics suggests that under a resource-constrained environment of 32 cores and 1 graphics card, the value model could squeeze out the last bit of performance, reaching 7.1\% (IMO-AG-30) and 3.4\% (MO-TG-225) improvement over a policy-only model.

\subsection*{Expert evaluation on TongGeometry results}

\begin{figure} 
    \centering
    \includegraphics[width=\textwidth]{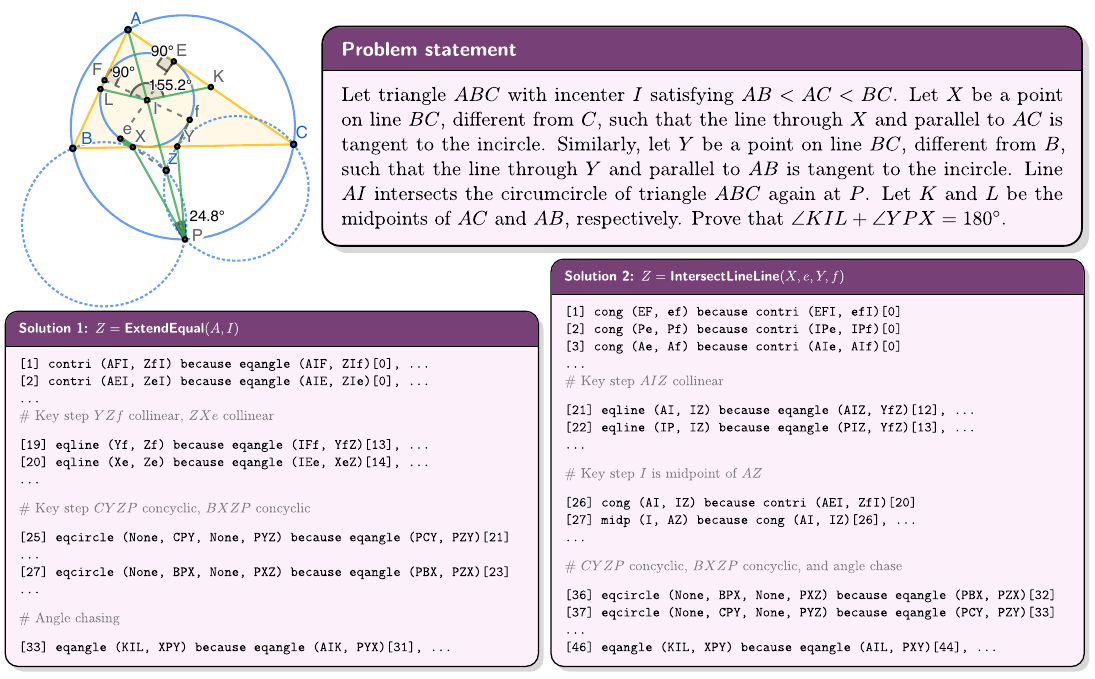} 

    \caption{\textbf{IMO 2024 P4 proof from TongGeometry.} TongGeometry generated two possible solutions to IMO 2024 P4, a problem that did not have well-documented solutions during TongGeometry training. TongGeometry constructed the same auxiliary point as the official solution. TongGeometry's proof was deemed correct by a 2024 IMO gold medalist. Since TongGeometry uses full angles, the original statement is equivalent to $\measuredangle KIL = \measuredangle XPY$.}
    \label{fig:proof} 
\end{figure}

Figure~\ref{fig:proof} shows IMO 2024 P4, the geometry problem in the latest IMO competition, a relatively new problem without many documented solutions at the time of TongGeometry training. TongGeometry generated two equivalent solutions. In retrospect, we observed that the two solutions introduced an auxiliary point identical to what's presented in the official solution~\cite{imo2024international}. We invited an 2024 IMO gold medalist to evaluate TongGeometry's solutions, and both were deemed correct.

\section*{Conclusion}

In this work, we introduce TongGeometry, a neuro-symbolic system designed to discover, present, and prove IMO-level geometry problems through guided tree search. We implemented evaluation rubrics for assessing the quality of TongGeometry-proposed problems. Notably, one proposed problem was selected in 2024 National High School Mathematics League (Beijing), a Team China qualifying competition, and two others shortlisted in 2024 US Ersatz Math Olympiad, a top civil math competition in the US. TongGeometry employs actor-critic style inference, where a policy model is learned to complete auxiliaries and a value model to estimate the number of remaining steps until problem solved. This style of problem solving leads to super-gold-medalist performance for the first time on IMO problems, surpasses the previous state-of-the-art, and demonstrates more effective use of neural models for auxiliary construction.

By analogy, TongGeometry functions more like a coach rather than a student merely solving problems --- a distinction that opens avenues for future study to further advance the field of geometry and mathematics.


\newpage


\begin{figure} 
    \centering
    \includegraphics[width=\textwidth]{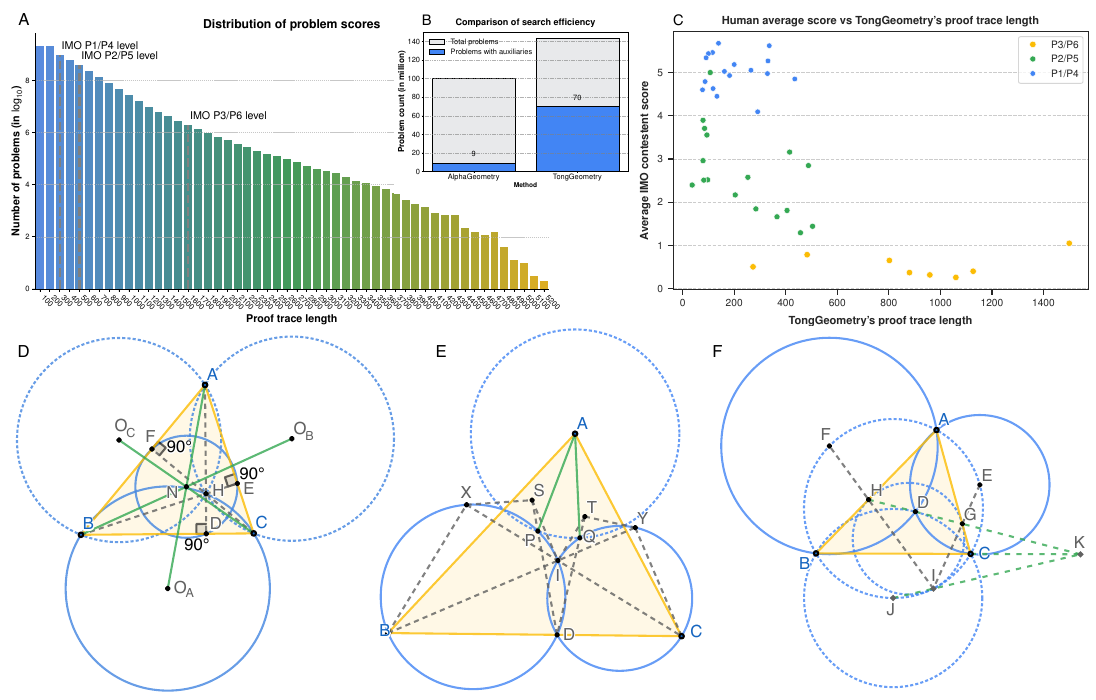} 

    \caption{\textbf{Generated data, statistics, and their distribution.} (A) The distribution of proof trace length in the generated problems. (B) Comparison of search efficiency of our method and AlphaGeometry. (C) The relationship of proof trace length and problem difficulty. (D) A known lemma discovered by TongGeometry. (E) A symmetric problem proposed by TongGeometry in 2024 USEMO shortlist that asks to prove AP and AQ are of the same length. (F) The Mixtilinear Incircle configuration found in TongGeometry's search buffer.}
    \label{fig:stats_examples} 
\end{figure}

\clearpage
\newpage


\begin{table}[t]
    \centering
    \caption{\textbf{Number of problems solved in IMO-AG-30 (left) and MO-AG-225 (right) in ascending order.} TongGeometry w/o value denotes TongGeometry without using the valued model trained with process-based reward learning.}
    \begin{tabular}{crr}
            \\
            \hline
            Method                 & Solves in IMO-AG-30 & Solves in MO-TG-225 \\
            \hline
            GPT-4o                 & 0 / 30              & 2 / 225             \\
            o1-preview             & 0 / 30              & 6 / 225             \\
            DD+AR (AlphaGeometry)  & 14 / 30             & 74 / 225            \\
            Honorable mentions     & 14.3 / 30           & -                   \\
            Average IMO contestant & 15.2 / 30           & -                   \\
            DD (TongGeometry)      & 18 / 30             & 101 / 225           \\
            Bronze medalist        & 19.3 / 30           & -                   \\
            Silver medalist        & 22.9 / 30           & -                   \\
            AlphaGeometry          & 25 / 30             & 102 / 225           \\
            Gold medalist          & 25.9 / 30           & -                   \\
            TongGeometry w/o value & 28 / 30             & 177 / 225           \\
            TongGeometry           & 30 / 30             & 183 / 225           \\
            \hline
    \end{tabular}
    \label{tab:perf}
\end{table}


\clearpage 

%
\bibliography{references} 
\bibliographystyle{sciencemag}

%
%
%
%
%
%


\section*{Acknowledgments}

This project is a two-year long collaboration between Beijing Institute for General Artificial Intelligence and Peking University. Along the journey, we would like to thank Jingyu Xie (Winter Camp) and Xingjian Liang (2023 IMO Gold Medalist) for assistance in problem examination, Patrik Bak (GeoGen author) for very thoughtful and detailed discussion for designing automatic rubrics and generating good proposals, Evan Chen (USA IMO coach) for assistance in problem evaluation in USEMO, Chen Zhen for crafting beautiful figures, Qing Li and Junqi Wang for early discussion, Liangru Xiang and Aoyang Qin for consistent assistance, and Xianbang Wang (2024 IMO Gold Medalist) for generated proof assessment.

\paragraph*{Funding:} C.~Z., J.~S., S.~L, Y.~L., Y.~M., W.~W., and S.-C.~Z. were supported in part by Beijing Institute for General Artificial Intelligence (BIGAI) (No. KY20220060). All authors were supported in part by the National Science and Technology Major Project (No. 2022ZD0114900). Y.~M. and Y.~Z. were supported in part by the National Natural Science Foundation of China (No. 62376031).


\paragraph*{Author contributions:}
Conceptualization: C.~Z., Y.~L.\\
Data curation: C.~Z.\\
Formal analysis: C.~Z., J.~S., S.~L.\\
Funding acquisition: Y.~Z., S.-C.~Z.\\
Investigation: C.~Z.\\
Methodology: C.~Z., Y.~L.\\
Project administration: C.~Z., W.~W., S.-C.~Z.\\
Software: C.~Z., J.~S.\\
Resources: Y.~M., Y.~Z., S.-C.~Z.\\
Supervision: W.~W., S.-C.~Z.\\
Validation: C.~Z.\\
Visualization: C.~Z.\\
Writing - original draft: C.~Z.\\
Writing - review \& editing: C.~Z., Y.~L, Y.~Z.

\paragraph*{Competing interests:} There are no competing interests to declare.


\newpage


\renewcommand{\thefigure}{S\arabic{figure}}
\renewcommand{\thetable}{S\arabic{table}}
\renewcommand{\theequation}{S\arabic{equation}}
\renewcommand{\thepage}{S\arabic{page}}
\setcounter{figure}{0}
\setcounter{table}{0}
\setcounter{equation}{0}
\setcounter{page}{1} 


\begin{center}
\section*{Supplementary Materials for\\ \scititle}

Chi~Zhang$^{\ast}$,
Jiajun~Song,
Siyu~Li,
Yitao~Liang,
Yuxi~Ma\\
Wei~Wang,
Yixin~Zhu$^{\ast}$,
Song-Chun~Zhu$^{\ast}$\\ 
\small$^\ast$Corresponding author. Email: zhangchi@bigai.ai, yixin.zhu@pku.edu.cn, sczhu@bigai.ai\\
\end{center}



\newpage


\subsection*{Supplementary Text}

\subsubsection*{Automated geometry theorem proposing and proving}

GeoGen~\cite{bak2020automated} is currently the only known program capable of automatically proposing olympiad geometry theorems. Like our approach, GeoGen begins with a blank canvas and randomly draws geometric elements. However, GeoGen is more accurately described as a theorem guesser rather than a theorem prover. While it can verify whether a fact holds numerically after constructing a diagram, it lacks the ability to construct auxiliary elements necessary for rigorous proof. As a result, the proving process relies heavily on human involvement. In contrast, TongGeometry not only proposes problems but also autonomously proves theorems.

Automatic geometry theorem proving methods can be broadly categorized into two approaches: algebraic methods and search-based technique. Wu's method~\cite{wu2008decision,chou1988mechanical} is the most prominent in the former line of research, in which a geometry theorem is formulated using a system of algebraic equations. The conclusion is considered proven if the system, after reduction~\cite{chou1984proving,chou1986proving}, has a zero remainder. While Wu's method can theoretically check the validity of any equality fact for geometry problems, there are two major concerns that hinder the wider adoption of Wu's method in geometry theorem proving. For one thing, Wu's method is purely algebraic and of limited interest to human readers. The proof process is mechanical, with reduction and equation solving carrying little semantic meaning corresponding to the diagrams, much as how contemporary mathematicians view analytical proving as less elegant. For another, Wu's method is inefficient in terms of space and time complexity, limiting its practical use.

With the development of deductive database~\cite{chou2000deductive}, area method~\cite{chou1994machine}, and full angles~\cite{chou1996automated}, the research community is more invested in the second, more human-like approach for geometry theorem proving. The Lean interactive theorem prover~\cite{moura2021lean} has also incorporated geometry libraries; yet these libraries based on either completed Euclid's axioms, Hilbert's axioms~\cite{hilbert1902foundations}, or Tarski's axioms~\cite{tarski1998decision} are very limited in their ability to prove olympiad-level problems. Besides, proof using axiomatic geometry is less intuitive and not as easily understandable as plain synthetic geometry. In addition, none of the methods properly address how the auxiliaries should be constructed to advance reasoning, other than using manually designed heuristics~\cite{matsuda2004gramy,wang2015automated}. AlphaGeometry~\cite{trinh2024solving} sidesteps this challenge using a data-driven method similar to ours. However, their data engine is inadequately designed to quantify the space of olympiad geometry, leaving it underexplored and resulting in performance that falls short of top-tier competition standards.

\subsubsection*{Learning-based methods for theorem proving}

The pursuit of automated theorem proving has its roots in the early development of logical systems. Building on this foundation, a range of powerful logic solvers, such as E~\cite{schulz2002brainiac}, SPASS~\cite{weidenbach2009spass}, and Z3~\cite{de2008z3}, have been developed. Recent advancements in interactive proof assistants, like Lean~\cite{moura2021lean}, have enabled the resolution of more complex theorems, facilitated by human guidance. Simultaneously, the AI community has been working towards minimizing human intervention in the theorem proving process by generating programmatic input, thus pushing the boundaries of automation. Following the breakthroughs in deep learning~\cite{lecun2015deep}, learning-based approaches have become the predominant focus of research in this area.

The deep learning era has ushered in two distinct streams of research in learning-based theorem proving: one focusing on learning-guided search and the other on improved data curation.

In learning-guided search, deep learning models serve as hint providers for the next steps taken by proof assistants~\cite{loos2017deep,wei2024proving}. Due to the absence of a direct mechanism to evaluate the utility of these actions, the log probability of the action is often employed as a heuristics~\cite{polu2020generative}. Utilizing best-first search or beam search, these systems have successfully solved a few simple olympiad problems of algebra and number theory~\cite{zheng2021minif2f} and certain geometry problems~\cite{trinh2024solving}. The success of learning a value function and its integration with a policy function in Go~\cite{silver2016mastering,silver2017mastering,silver2018general} have inspired the community to explore learning a utility function for theorem proving actions. For instance, HyperTree~\cite{lample2022hypertree} employs AlphaZero-style Monte Carlo Tree Search to learn both a policy model and a critics model to guide the proof search.

High-quality data is considered a crucial element for the success of learning-based methods. However, in contrast to other domains, mathematical data is substantially more scarce. The most widely used datasets, such as Metamath~\cite{megill2019metamath} and Mathlib~\cite{moura2021lean}, contain fewer than 60,000 human-written theorems combined. The miniF2F benchmark~\cite{zheng2021minif2f,polu2022formal} includes only a few hundred validation and test problems. Quality mathematical data is difficult to find naturally; for instance, the IMO produces just six problems per year. This scarcity has led researchers to explore synthetic data as an alternative. Models trained on data generated from specially designed symbolic systems have demonstrated impressive performance~\cite{aygun2022proving,firoiu2021training,trinh2024solving}. Additionally, advancements in LLMs have prompted investigations into their use as data generators~\cite{liu2023tinygsm}. Informal proofs have also been used to augment data via autoformalization~\cite{wu2022autoformalization}. An emerging technique in data curation is process-based supervision~\cite{uesato2022solving,lightman2023let}, where each step of the proof is supervised. However, this approach requires more extensive annotation efforts compared to outcome-based supervision.

\subsubsection*{Explanation on the Nine-point Center}

In Figure~\ref{fig:stats_examples}(D), in $\triangle ABC$, $D$, $E$, and $F$ are the feet corresponding to $A$, $B$, and $C$, respectively. Let $H$ be the orthocenter of $\triangle ABC$ and $O_A$ be the circumcenter of $\triangle BCH$. TongGeometry finds that the circumcenter $N$ of $\triangle DEF$ happens to be the midpoint of $AO_A$. The fact builds on several important theorems in geometry literature: the circumcircle of $\triangle DEF$ is the famous Nine-point circle~\cite{mackay1892history} and the circumcenter of $\triangle ABC$ and the nine-point circle center of $\triangle ABC$ are homothetic with respect to its orthocenter in a ratio of 1:2~\cite{chen2021euclidean}. Via angle chasing, one can prove $O_A$ and the circumcenter of $\triangle ABC$ are symmetric with respect to $BC$, hence the result. Note that the original problem is mirror-symmetric in that $B-C$ and $E-F$ are symmetric pairs. In addition, the problem can be made rotationally symmetric, by constructing circumcircles of $\triangle ACH$ and $\triangle ABH$ and claiming that $N$ is the midpoint of $AO_A$, $BO_B$ and $CO_C$.

\subsubsection*{Explanation on the Mixtilinear Incircle}

In Figure~\ref{fig:stats_examples}(F), the base configuration is described as follows: in $\triangle ABC$, let $D$ be its incenter, and $E$ and $F$ the circumcenters of $\triangle ACD$ and $\triangle ABD$, respectively. Points $H$ and $G$ lie on $AB$ and $AC$, respectively, such that $H$, $D$, and $G$ are collinear and perpendicular to $AD$. This foundational setup reveals multiple intrinsic geometric properties. Notably, $ABCEF$ are concyclic, with $E$ and $F$ marking the arc midpoints. The intersection of $EG$ and $FH$, labeled $I$, forms a circle with $H$ and $G$, which is tangent to $AB$, $AC$, and $\odot(ABC)$ at $H$, $G$, and $I$, respectively. Denoting the circumcenter of $\triangle BCD$ as $J$, the lines $HG$, $BC$, and $IJ$ are concurrent at a point $K$.

\subsubsection*{Limitation}

Detailed analysis suggests that current models still fall short in sequentially constructing a series of interrelated actions, failing on problems requiring multiple new points to be constructed. Such a problem could be rooted in the data-driven nature of neural models: the longer the number of steps to take, the (exponentially) fewer the data points. Additionally, a disconnect exists between the natural language representation of geometry problems and the domain-specific languages used in computational settings. While the computer programs use direct sequential construction, how to correctly convert natural language representation to computer construction remains an under-explored issue. Human-proposed problems frequently obscure straightforward constructions with constraints, introducing intentional complexities that contribute to the problem's difficulty.

\subsubsection*{IMO 2024 P4 Full Solutions}

\begingroup
\scriptsize
\begin{verbatim}
Z = ExtendEqual(A, I)
[1] contri (AFI, ZfI) because eqangle (AIF, ZIf)[0], cong (FI, If)[0], cong (AI, IZ)[0]
[2] contri (AEI, ZeI) because eqangle (AIE, ZIe)[0], cong (EI, Ie)[0], cong (AI, IZ)[0]
[3] simtri (EFe, ZfI) because eqangle (EeF, ZIf)[0], eqratio (Ee, IZ, Fe, If)[0]
[4] simtri (EFe, ZeI) because eqangle (EeF, eIZ)[0], eqratio (Ee, IZ, Fe, Ie)[0]
[5] simtri (ACZ, AKI) because eqangle (CAZ, KAI)[0], eqratio (AC, AK, AZ, AI)[0]
[6] simtri (ABZ, ALI) because eqangle (BAZ, LAI)[0], eqratio (AB, AL, AZ, AI)[0]
[7] eqangle (AFI, ZfI) because contri (AFI, ZfI)[1]
[8] eqangle (AEI, ZeI) because contri (AEI, ZeI)[2]
[9] eqangle (FEe, fZI) because simtri (EFe, ZfI)[3]
[10] eqangle (FEe, IZe) because simtri (EFe, ZeI)[4]
[11] eqangle (AIK, AZC) because simtri (ACZ, AKI)[5]
[12] eqangle (AIL, AZB) because simtri (ABZ, ALI)[6]
[13] eqangle (IFf, YfZ) because eqangle (AFI, ZfI)[7], eqangle (AFf, YfI)[0]
[14] eqangle (IEe, XeZ) because eqangle (AEI, ZeI)[8], eqangle (AEe, XeI)[0]
[15] eqangle (AZf, PCY) because eqangle (FEe, YCP)[0], eqangle (FEe, fZI)[9], eqangle (AZf, IZf)[0]
[16] eqangle (AZe, PBX) because eqangle (AZe, IZe)[0], eqangle (FEe, IZe)[10], eqangle (FEe, PBX)[0]
[17] eqangle (AIK, PZC) because eqangle (AIK, AZC)[11], eqangle (AZC, PZC)[0]
[18] eqangle (AIL, PZB) because eqangle (AIL, AZB)[12], eqangle (AZB, PZB)[0]
[19] eqline (Yf, Zf) because eqangle (IFf, YfZ)[13], eqline (FI, Ff)[0]
[20] eqline (Xe, Ze) because eqangle (IEe, XeZ)[14], eqline (EI, Ee)[0]
[21] eqangle (PCY, PZY) because eqangle (AZY, PZY)[0], eqangle (AZY, AZf)[19], eqangle (AZf, PCY)[15]
[22] eqangle (CYf, XYZ) because eqangle (AYZ, AYf)[19], eqangle (AYC, AYX)[0]
[23] eqangle (PBX, PZX) because eqangle (AZX, PZX)[0], eqangle (AZX, AZe)[20], eqangle (AZe, PBX)[16]
[24] eqangle (BXe, YXZ) because eqangle (AXZ, AXe)[20], eqangle (AXB, AXY)[0]
[25] eqcircle (None, CPY, None, PYZ) because eqangle (PCY, PZY)[21]
[26] eqangle (CYP, XYP) because eqangle (PYZ, PYf)[19], eqangle (CYf, XYZ)[22]
[27] eqcircle (None, BPX, None, PXZ) because eqangle (PBX, PZX)[23]
[28] eqangle (BXP, YXP) because eqangle (PXZ, PXe)[20], eqangle (BXe, YXZ)[24]
[29] eqangle (CYP, CZP) because eqcircle (None, CPY, None, PYZ)[25]
[30] eqangle (BXP, BZP) because eqcircle (None, BPX, None, PXZ)[27]
[31] eqangle (AIK, PYX) because eqangle (AIK, PZC)[17], eqangle (CYP, CZP)[29], eqangle (CYP, XYP)[26]
[32] eqangle (AIL, PXY) because eqangle (AIL, PZB)[18], eqangle (BXP, BZP)[30], eqangle (BXP, YXP)[28]
[33] eqangle (KIL, XPY) because eqangle (AIK, PYX)[31], eqangle (AIL, PXY)[32]
\end{verbatim}

\begin{verbatim}
Z = IntersectLineLine(X, e, Y, f)
[1] cong (EF, ef) because contri (EFI, efI)[0]
[2] cong (Pe, Pf) because contri (IPe, IPf)[0]
[3] cong (Ae, Af) because contri (AIe, AIf)[0]
[4] simtri (AEF, Zfe) because eqangle (AFE, feZ)[0], eqangle (AEF, efZ)[0]
[5] eqcircle (None, IZe, None, IZf) because eqangle (IeZ, IfZ)[0]
[6] eqratio (AE, Zf, EF, ef) because simtri (AEF, Zfe)[4]
[7] eqangle (ZIe, Zfe) because eqcircle (None, IZe, None, IZf)[5]
[8] eqangle (IZf, Ief) because eqcircle (None, IZe, None, IZf)[5]
[9] contri (PZe, PZf) because cong (Ze, Zf)[0], cong (Pe, Pf)[2]
[10] contri (AZe, AZf) because cong (Ze, Zf)[0], cong (Ae, Af)[3]
[11] cong (AE, Zf) because eqratio (AE, Zf, EF, ef)[6], cong (EF, ef)[1]
[12] eqangle (AIZ, YfZ) because eqangle (ZIe, Zfe)[7], eqangle (AIe, Yfe)[0]
[13] eqangle (PIZ, YfZ) because eqangle (ZIe, Zfe)[7], eqangle (PIe, Yfe)[0]
[14] eqangle (IAL, IZY) because eqangle (IAL, Ief)[0], eqangle (IZf, Ief)[8], eqangle (IZY, IZf)[0]
[15] eqangle (IZY, XBP) because eqangle (Ief, XBP)[0], eqangle (IZf, Ief)[8], eqangle (IZY, IZf)[0]
[16] eqangle (IAK, YZI) because eqangle (IAK, feI)[0], eqangle (IZf, Ief)[8], eqangle (IZY, IZf)[0]
[17] eqangle (IZY, PCY) because eqangle (Ief, PCY)[0], eqangle (IZf, Ief)[8], eqangle (IZY, IZf)[0]
[18] eqangle (PZe, fZP) because contri (PZe, PZf)[9]
[19] eqangle (AZe, fZA) because contri (AZe, AZf)[10]
[20] contri (AEI, ZfI) because cong (AE, Zf)[11], perp (AEI)[0], perp (IfZ)[0], cong (EI, If)[0]
[21] eqline (AI, IZ) because eqangle (AIZ, YfZ)[12], eqline (Yf, Zf)[0]
[22] eqline (IP, IZ) because eqangle (PIZ, YfZ)[13], eqline (Yf, Zf)[0]
[23] eqangle (PZX, YZP) because eqangle (PZX, PZe)[0], eqangle (PZe, fZP)[18], eqangle (PZY, PZf)[0]
[24] eqangle (AZP, PZA) because eqangle (PZe, fZP)[18], eqangle (AZe, fZA)[19]
[25] eqangle (AZY, CAZ) because eqangle (AZe, ZAC)[0], eqangle (AZe, fZA)[19], eqangle (AZY, AZf)[0]
[26] cong (AI, IZ) because contri (AEI, ZfI)[20]
[27] midp (I, AZ) because cong (AI, IZ)[26], eqline (AI, IZ)[21]
[28] eqratio (AB, AL, AZ, AI) because midp (I, AZ)[27], midp (L, AB)[0]
[29] eqratio (AC, AK, AZ, AI) because midp (I, AZ)[27], midp (K, AC)[0]
[30] eqangle (BAZ, LAI) because eqangle (AZY, ZAB)[0], eqangle (AZY, IZY)[21], eqangle (IAL, IZY)[14]
[31] eqangle (CAZ, KAI) because eqangle (AZY, CAZ)[25], eqangle (AZY, IZY)[21], eqangle (IAK, YZI)[16]
[32] eqangle (PBX, PZX) because eqangle (IZY, XBP)[15], eqangle (IZY, PZY)[22], eqangle (PZX, YZP)[23]
[33] eqangle (PCY, PZY) because eqangle (IZY, PCY)[17], eqangle (IZY, PZY)[22]
[34] simtri (ABZ, ALI) because eqratio (AB, AL, AZ, AI)[28], eqangle (BAZ, LAI)[30]
[35] simtri (ACZ, AKI) because eqratio (AC, AK, AZ, AI)[29], eqangle (CAZ, KAI)[31]
[36] eqcircle (None, BPX, None, PXZ) because eqangle (PBX, PZX)[32]
[37] eqcircle (None, CPY, None, PYZ) because eqangle (PCY, PZY)[33]
[38] eqangle (AIL, AZB) because simtri (ABZ, ALI)[34]
[39] eqangle (AIK, AZC) because simtri (ACZ, AKI)[35]
[40] eqangle (BXP, BZP) because eqcircle (None, BPX, None, PXZ)[36]
[41] eqangle (CYP, CZP) because eqcircle (None, CPY, None, PYZ)[37]
[42] eqangle (AZB, PXY) because eqangle (AZB, PZB)[24], eqangle (BXP, BZP)[40], eqangle (BXP, YXP)[0]
[43] eqangle (AZC, PYX) because eqangle (AZC, PZC)[24], eqangle (CYP, CZP)[41], eqangle (CYP, XYP)[0]
[44] eqangle (AIL, PXY) because eqangle (AIL, AZB)[38], eqangle (AZB, PXY)[42]
[45] eqangle (AIK, PYX) because eqangle (AIK, AZC)[39], eqangle (AZC, PYX)[43]
[46] eqangle (KIL, XPY) because eqangle (AIL, PXY)[44], eqangle (AIK, PYX)[45]
\end{verbatim}
\endgroup

\subsubsection*{TongGeometry Proposals to 2024 USEMO}

\begin{figure}
    \centering
    \includegraphics{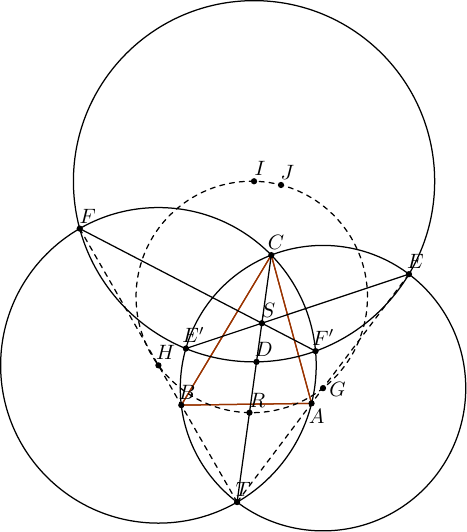}
    \caption{\textbf{Problem 1 of TongGeometry's proposal to 2024 USEMO.}}
    \label{fig:sup_proposal_usemo_1}
\end{figure}

\begin{figure}
    \centering
    \includegraphics{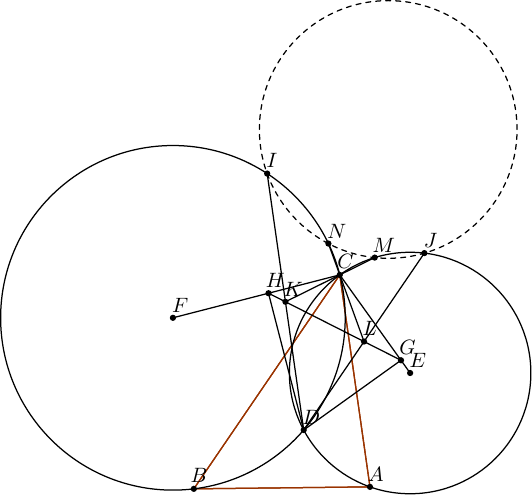}
    \caption{\textbf{Problem 2 of TongGeometry's proposal to 2024 USEMO.}}
    \label{fig:sup_proposal_usemo_2}
\end{figure}

\begin{figure}
    \centering
    \includegraphics{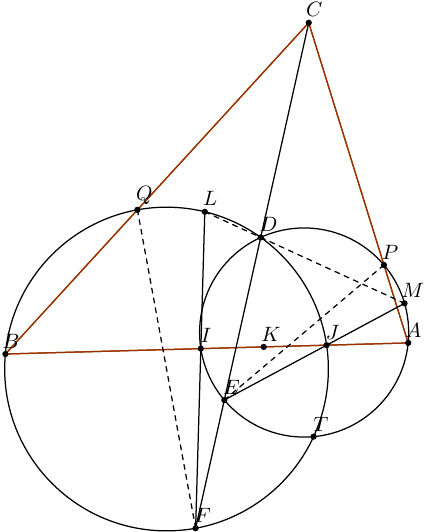}
    \caption{\textbf{Problem 3 of TongGeometry's proposal to 2024 USEMO.}}
    \label{fig:sup_proposal_usemo_3}
\end{figure}

\begin{figure}
    \centering
    \includegraphics{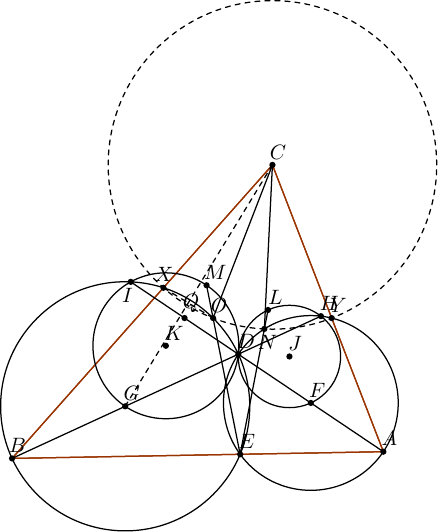}
    \caption{\textbf{Problem 4 of TongGeometry's proposal to 2024 USEMO.}}
    \label{fig:sup_proposal_usemo_4}
\end{figure}

\begin{figure}
    \centering
    \includegraphics{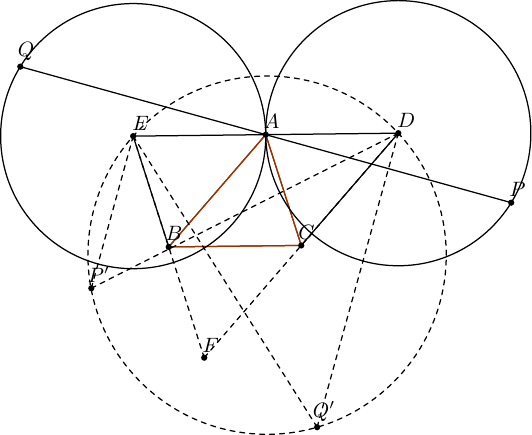}
    \caption{\textbf{Problem 5 of TongGeometry's proposal to 2024 USEMO.}}
    \label{fig:sup_proposal_usemo_5}
\end{figure}

\begin{figure}
    \centering
    \includegraphics{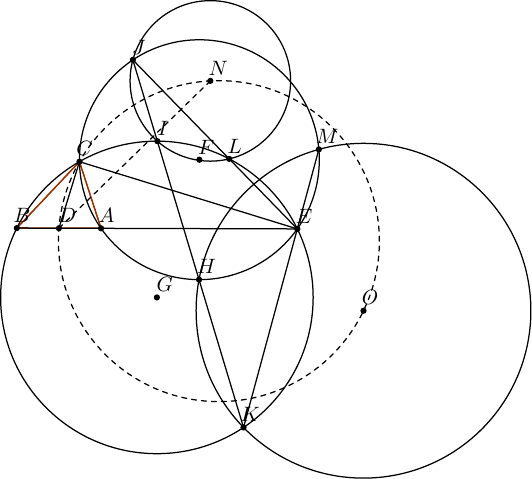}
    \caption{\textbf{Problem 6 of TongGeometry's proposal to 2024 USEMO.}}
    \label{fig:sup_proposal_usemo_6}
\end{figure}

\begin{enumerate}
\item 
Figure~\ref{fig:sup_proposal_usemo_1}: Let $\triangle ABC$ be an acute-angled triangle, $D$ be its incenter, $E$ be its $B$-excenter and $F$ its $A$-excenter. $G$ is the center of $\odot(CBE)$ and $H$ is the center of $\odot(ACF)$. A circumcircle centered at $I$ passes through $\triangle DEF$. $J$ is the foot of $I$ to $CD$. Prove $GHIJ$ are concyclic.

\item 
Figure~\ref{fig:sup_proposal_usemo_2}: Let $\triangle ABC$ be an acute-angled triangle and $D$ be its incenter. Let $E$ be the circumcenter of $\odot(ACD)$ and $F$ be that of $\odot(BCD)$. Project $D$ to $CE$ to get $G$ and project $D$ to $CF$ to get $H$. Denote the other intersection of $EC$ with $\odot(BCD)$ as $I$ and that of $FC$ with $\odot(ACD)$ as $J$. $ID$ and $JD$ intersect with $HG$ at $K$ and $L$, respectively. Prove $I$ and $J$ and the other intersection of $LC$ with $\odot(CDB)$ and that of $KC$ with $\odot(DCA)$ concyclic.

\item 
Figure~\ref{fig:sup_proposal_usemo_3}: Let $\triangle ABC$ be an acute-angled triangle and $D$ be its incenter. $E$ is on $CD$ such that $\angle EAC$ is a right angle. $F$ is on $CD$ such that $\angle FBC$ is a right angle. The circumcircle of $\triangle AED$ meets $BC$ again at $I$ and the circumcircle of $\triangle BFD$ meets $BC$ again at $J$. Let $K$ be the midpoint of $IJ$. Denote the second intersection of $IF$ with $\odot(BFD)$ and that of $EJ$ with $\odot(AED)$ as $L$ and $M$, respectively. Prove $KL = KM$.

\item 
Figure~\ref{fig:sup_proposal_usemo_4} (\textbf{shortlisted}): Let $\triangle ABC$ be an acute-angled triangle and $D$ be its incenter. The projection of $D$ on $AB$, denoted as $E$, forms two circumcircles with $AD$ and $BD$, centered at $F$ and $G$ respectively. $AD$ intersects $\odot(BDE)$ again at $I$ and $BD$ intersects $\odot(ADE)$ again at $H$. Denote the antipode of $G$ in $\odot(GID)$ as $M$ and that of $F$ as $L$. Prove $ME$ and $LE$ intersect $\odot(BDE)$ and $\odot(ADE)$ at the same distance from $C$.

\item 
Figure~\ref{fig:sup_proposal_usemo_5} (\textbf{shortlisted}): Let $\triangle ABC$ be a scalene triangle, $ABCD$ and $ACBE$ are two parallelograms. Circle $E$ and $D$ pass $A$. An arbitrary line through $A$ meets the two circles again at $Q$ and $P$. Let $Q^\prime$ and $P^\prime$ denote the reflections of $Q$ against $B$ and $P$ against $C$, respectively. $F$ being the intersection of $DC$ and $EB$, what conditions should $PQ$ and $AF$ satisfy to make $DEQ^\prime P^\prime$ concyclic?

\item 
Figure~\ref{fig:sup_proposal_usemo_6}: Let $\triangle ABC$ be an acute-angled triangle and $D$ be $AB$'s midpoint. $E$ is on line $AB$ and $CD \perp CE$. The line linking the midpoint $H$ of minor arc $AE$ in $\odot(ACE)$ and the midpoint $I$ of minor arc $BE$ in $\odot(BCE)$ intersects $\odot(ACE)$ again at $J$ and $\odot(BCE)$ again at $K$. $JE$ meets $\odot(BCE)$ again at $L$ and $KE$ meets $\odot(ACE)$ at $M$. Prove the circumcenter of $\odot(IJL)$, the circumcenter of $\odot(HKM)$ and $C$ and $D$ are concyclic.
\end{enumerate}

\subsubsection*{TongGeometry Proposals to 2024 NHSML (Beijing)}

\begin{figure}
    \centering
    \includegraphics{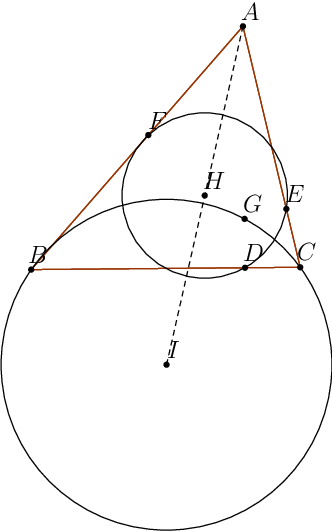}
    \caption{\textbf{Problem 1 of TongGeometry's proposal to 2024 NHSML (Beijing).}}
    \label{fig:sup_proposal_nhsml_1}
\end{figure}

\begin{figure}
    \centering
    \includegraphics{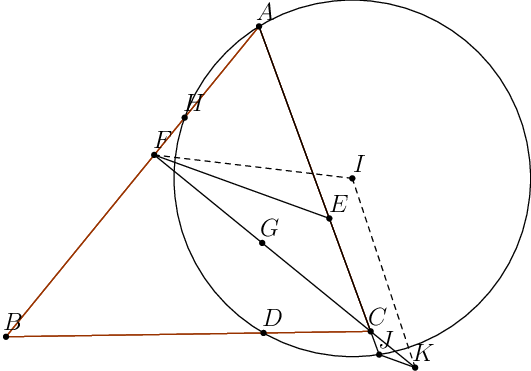}
    \caption{\textbf{Problem 2 of TongGeometry's proposal to 2024 NHSML (Beijing).}}
    \label{fig:sup_proposal_nhsml_2}
\end{figure}

\begin{figure}
    \centering
    \includegraphics{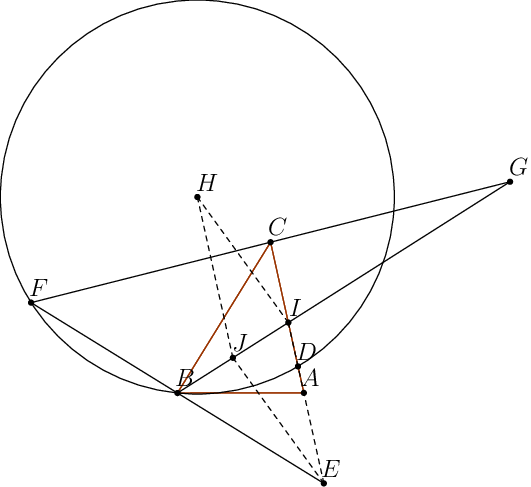}
    \caption{\textbf{Problem 3 of TongGeometry's proposal to 2024 NHSML (Beijing).}}
    \label{fig:sup_proposal_nhsml_3}
\end{figure}

\begin{figure}
    \centering
    \includegraphics{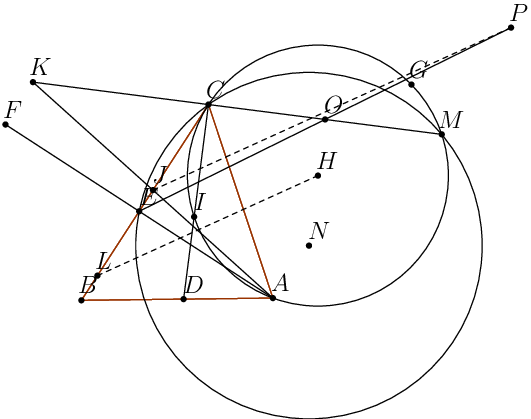}
    \caption{\textbf{Problem 4 of TongGeometry's proposal to 2024 NHSML (Beijing).}}
    \label{fig:sup_proposal_nhsml_4}
\end{figure}

\begin{enumerate}
\item 
Figure~\ref{fig:sup_proposal_nhsml_1}: Let $\triangle ABC$ be an acute-angled triangle, $D$, $E$, $F$ be feet corresponding to $A$, $B$, $C$, respectively. $G$ is $\triangle ABC$'s orthocenter. $H$ is the circumcenter of $\odot(DEF)$. $I$ is the circumcenter of $\odot(BCG)$. Prove $H$ is the mid point of $AI$.

\item 
Figure~\ref{fig:sup_proposal_nhsml_2} (\textbf{selected}): Let $\triangle ABC$ be an acute-angled triangle, $D$, $E$, $F$ be feet corresponding to $A$, $B$, $C$, respectively. $G$ is $\triangle ABC$'s orthocenter. $H$ is a point on $AB$ such that $DH$ is parallel to $AC$. $I$ is the circumcenter of $\odot(ADH)$. $J$ is another intersection of $AC$ and $\odot(I)$. $K$ is the intersection of $CF$ and a line through $J$ parallel to $EF$. Prove $IF = IK$.

\item 
Figure~\ref{fig:sup_proposal_nhsml_3}: Let $\triangle ABC$ be an acute-angled triangle, $D$ be $B$'s foot. $E$ is on $AC$ such that $\angle EBC$ is a right angle. $F$ is the reflection of $E$ over $B$. $G$ is the reflection of $F$ over $C$. Denote the intersection of $BG$ and $AC$ as $I$ and the mid point of $BI$ as $J$. $H$ is the circumcenter of $\odot(BDF)$. Prove $HIEJ$ is a parallelogram.

\item 
Figure~\ref{fig:sup_proposal_nhsml_4}: Let $\triangle ABC$ be an acute-angled triangle and $D$ the intersection of $AB$ and $\angle C$'s internal angle bisector. $E$ is the projection of $A$ on $BC$. $F$ is the reflection of $A$ over $E$. $G$ the reflection of $F$ over $C$. $H$ is the circumcenter of $\odot(ACG)$. $I$ is another intersection of $CD$ with $\odot(H)$. $J$ is the projection of $I$ on $BC$. $K$ is the reflection of $A$ over $J$. $L$ the reflection of $C$ over $J$. $KC$ intersects with $\odot(H)$ at $M$. $O$ is the mid point of $CM$. $P$ is the reflection of $E$ over $O$. Prove $HL$ is parallel to $JP$.
\end{enumerate}

\subsubsection*{Proof of Proposal Acceptance}

\begin{figure}
    \centering
    \includegraphics[width=\textwidth]{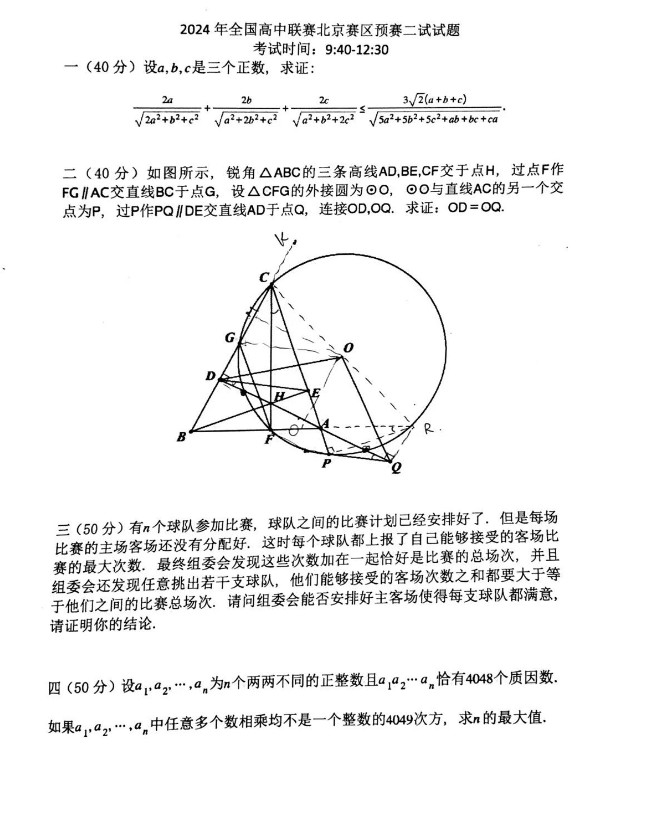}
    \caption{\textbf{Our proposal in 2024 NHSML (Beijing).} The second problem is our proposal, the only geometry in the competition.}
    \label{fig:supp_ppa_nhsml}
\end{figure}

Figure~\ref{fig:supp_ppa_nhsml} is the actual test in 2024 NHSML (Beijing). The second problem is our proposal.

\begin{figure}
    \centering
    \includegraphics[width=\textwidth]{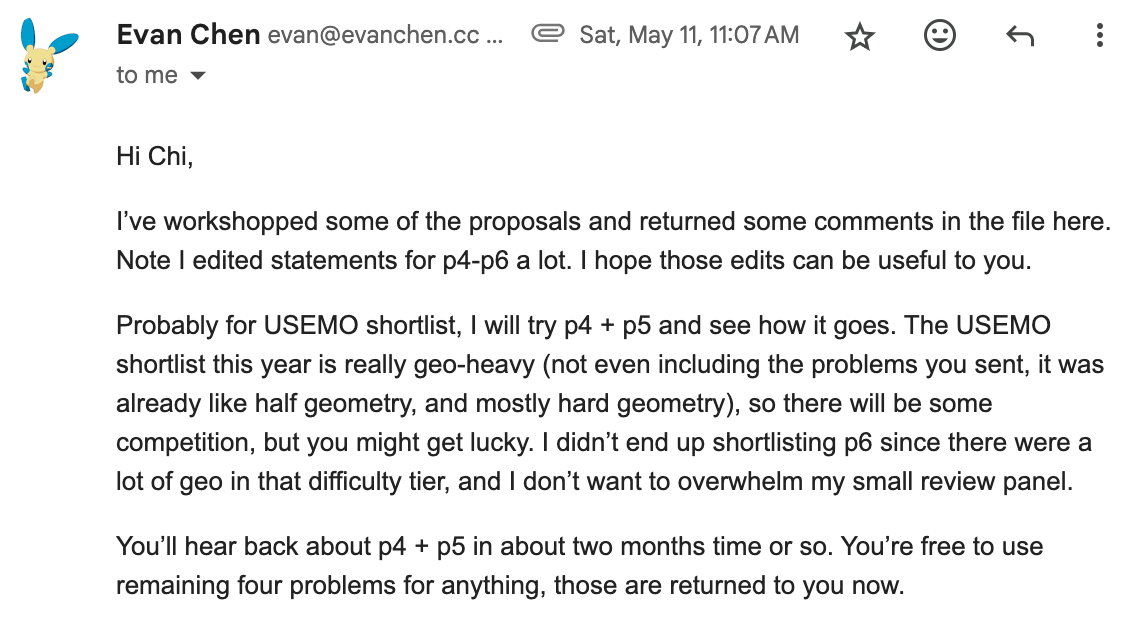}
    \caption{\textbf{Receipt of our proposals to 2024 USEMO.} The receipt confirms shortlisting two of our proposals.}
    \label{fig:supp_ppa_usemo}
\end{figure}

Figure~\ref{fig:supp_ppa_usemo} is the receipt from the organizer of 2024 USEMO confirming shortlisting our proposals.

\newpage




\begin{figure}
    \centering
    \includegraphics[width=\textwidth]{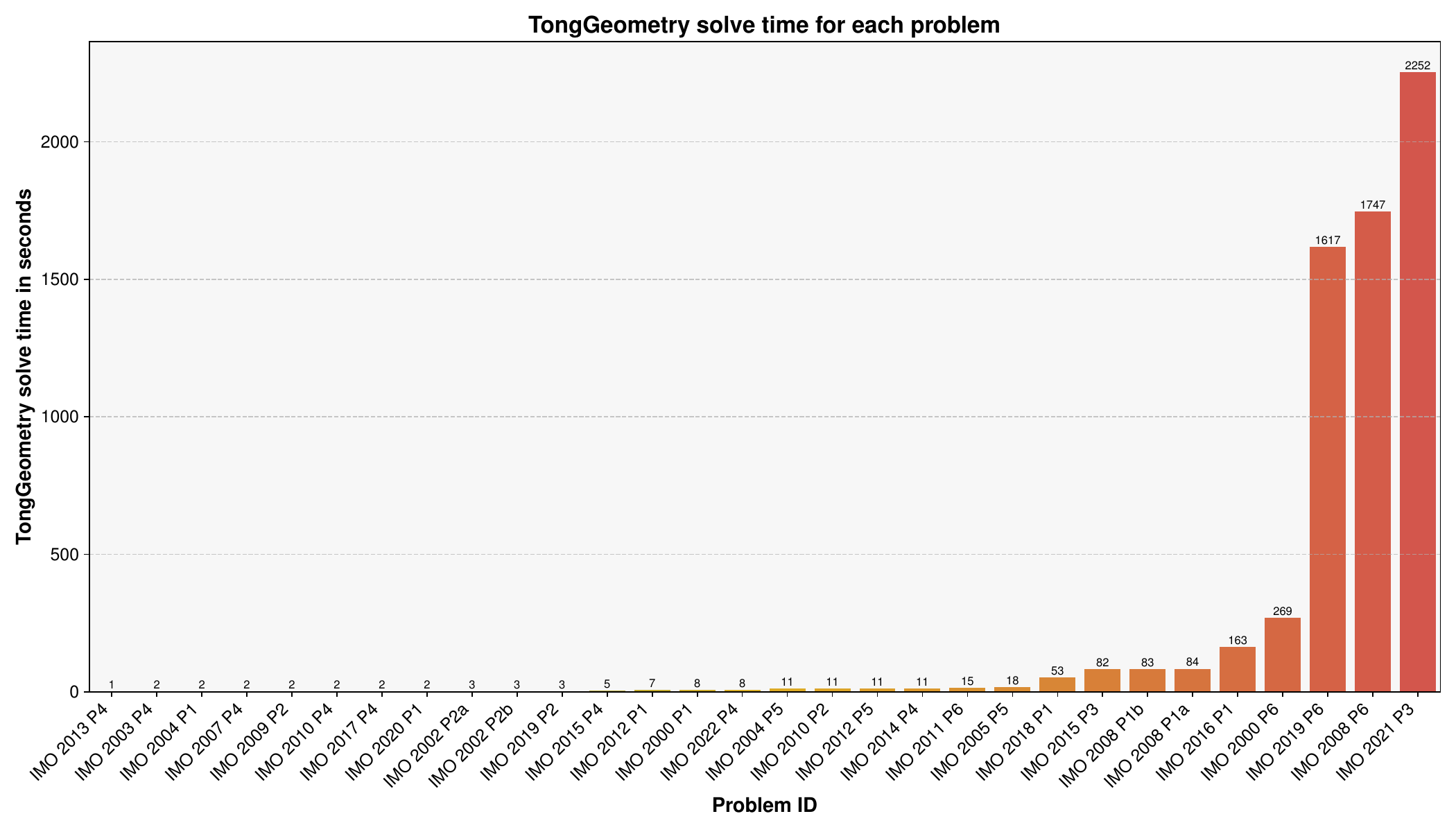}
    \caption{\textbf{TongGeometry's IMO problem solve time with 32 Intel Core i9-13900KF CPU cores and 1 NVIDIA RTX 4090 graphics card.} The solve time is denoted in second. Note that the longest solve time is within 38 minutes.}
    \label{fig:sup_time}
\end{figure}

\clearpage
\newpage



\clearpage 





\end{document}